\documentclass[10pt,twocolumn,letterpaper]{article}

\usepackage{cvpr}              

\usepackage[accsupp]{axessibility}

\definecolor{cvprblue}{rgb}{0.21,0.49,0.74}
\usepackage{amssymb}
\usepackage[pagebackref,breaklinks,colorlinks,allcolors=cvprblue]{hyperref}

\def\paperID{1334} 
\def\confName{CVPR}
\def\confYear{2025}

\title{The Devil is in the Prompts: Retrieval-Augmented Prompt Optimization for Text-to-Video Generation}

\author{
Bingjie Gao$^{1,2}$ \enspace\enspace\enspace\enspace\enspace
Xinyu Gao$^{1,2}$ \enspace\enspace\enspace\enspace\enspace 
Xiaoxue Wu$^{3,2}$ \enspace\enspace\enspace\enspace\enspace 
Yujie Zhou$^{1,2}$ \enspace \\
Yu Qiao$^{2}$\footnotemark[2]~~~~ \enspace 
Li Niu$^{1}$\footnotemark[2]~~~~ \enspace 
Xinyuan Chen$^{2}$\footnotemark[2]~~~~ \enspace 
Yaohui Wang$^{2}$\footnotemark[2] \enspace \\
\textsuperscript{1}Shanghai Jiao Tong University \enspace
\textsuperscript{2}Shanghai Artificial Intelligence Laboratory \enspace 
\textsuperscript{3}Fudan University \enspace 
}

\begin{document}
\maketitle
{
\renewcommand{\thefootnote}{\fnsymbol{footnote}}
\footnotetext[2]{Corresponding author}
}

\begin{figure*}[tb]
    \centering
\includegraphics[width=1.0\textwidth]{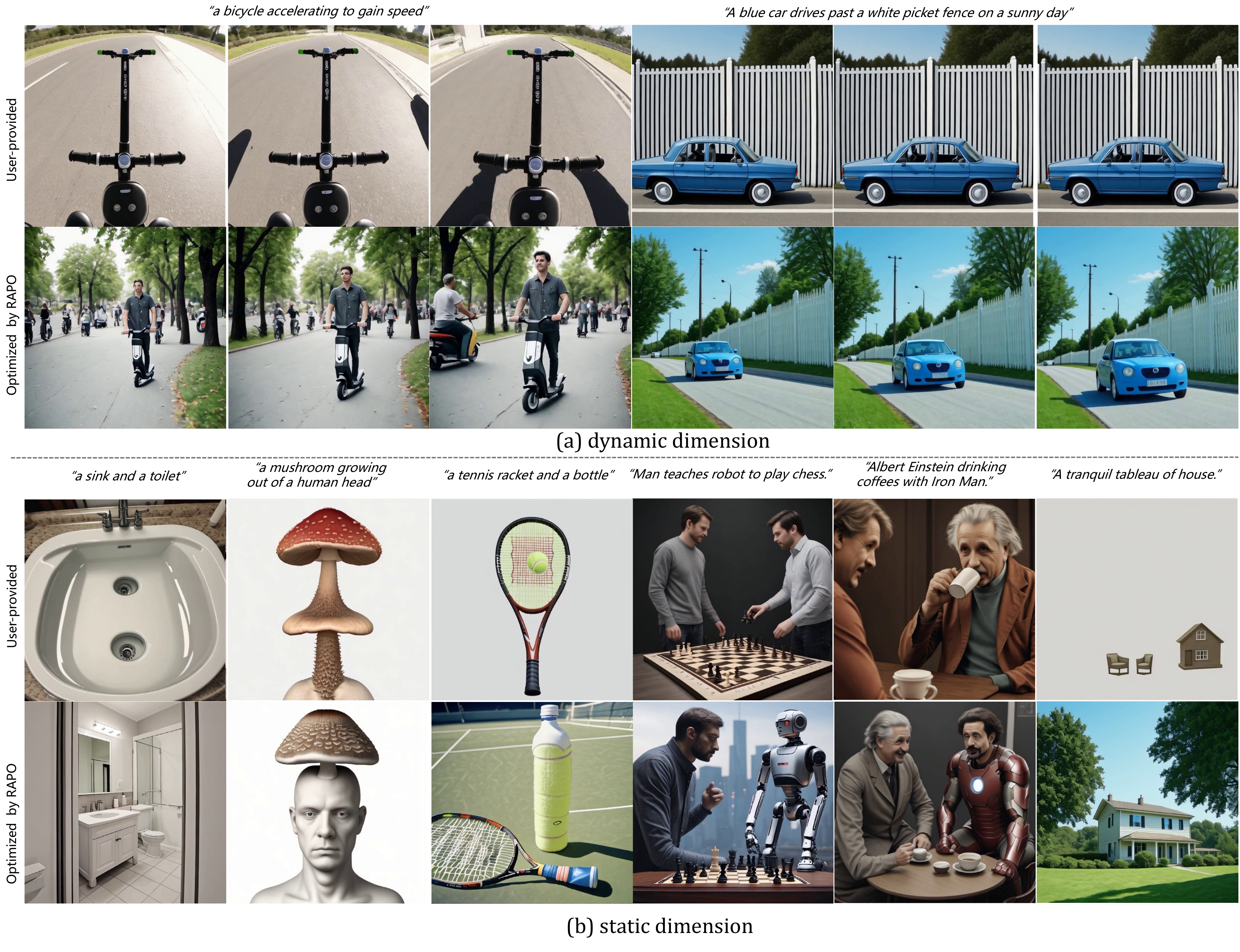}
    \caption{Videos generated using LaVie \cite{wang2023lavie} conditioned on user-provided prompts and optimized prompts from RAPO. The optimized prompts can significantly enhance both the static and dynamic qualities of the generated videos, making them more visually appealing.}
    \label{fig:teaser}
\end{figure*}
 
\begin{abstract}
The evolution of Text-to-video (T2V) generative models, trained on large-scale datasets, has been marked by significant progress. However, the sensitivity of T2V generative models to input prompts highlights the critical role of prompt design in influencing generative outcomes. Prior research has predominantly relied on Large Language Models (LLMs) to align user-provided prompts with the distribution of training prompts, albeit without tailored guidance encompassing prompt vocabulary and sentence structure nuances. To this end, we introduce \textbf{RAPO}, a novel \textbf{R}etrieval-\textbf{A}ugmented \textbf{P}rompt \textbf{O}ptimization framework. In order to address potential inaccuracies and ambiguous details generated by LLM-generated prompts. RAPO refines the naive prompts through dual optimization branches, selecting the superior prompt for T2V generation. The first branch augments user prompts with diverse modifiers extracted from a learned relational graph, refining them to align with the format of training prompts via a fine-tuned LLM. Conversely, the second branch rewrites the naive prompt using a pre-trained LLM following a well-defined instruction set. Extensive experiments demonstrate that RAPO can effectively enhance both the static and dynamic dimensions of generated videos, demonstrating the significance of prompt optimization for user-provided prompts. Project website: \href{https://whynothaha.github.io/Prompt_optimizer/RAPO.html}{GitHub}.
\end{abstract}    
\section{Introduction}
\label{sec:intro}

With the rapid advancement of diffusion models \cite{peebles2023scalable,ramesh2022hierarchical,rombach2022high}, visual content creation has experienced remarkable progress in recent years. The generation of images, as well as videos from text prompts utilizing large-scale diffusion models, referred to as text-to-images (T2I) \cite{podell2023sdxl,esser2024scaling,saharia2022photorealistic} and text-to-videos  (T2V) \cite{brooks2024video,esser2023structure,wang2023lavie} generation, have attracted significant interest due to the broad range of applications in real-world scenarios. Various efforts have been made to enhance the performance of these models, including improvements in model architecture \cite{jin2024pyramidal,ma2024latte,ma2024latte}, learning strategies \cite{yang2024cogvideox,singer2022make}, and data curation \cite{qiu2023freenoise,wang2023gen,wang2024loong}.

Recent studies \cite{polyak2024movie,hao2024optimizing,yang2024cogvideox} have revealed that employing long, detailed prompts with a pre-trained model typically produces superior quality outcomes compared to utilizing shallow descriptions provided by users. This has underscored the significance of prompt optimization as an important challenge in text-based visual content creation. The prompts provided by users are often brief and lack the essential details required to generate vivid images or videos. Simply attempting to optimize prompts by manually adding random descriptions can potentially mislead models and degrade the quality of generative results, resulting in outputs that may not align with user intentions. Therefore, developing automated methods to enhance user-provided prompts becomes essential for improving the overall quality of generated content.

Towards improving image aesthetics and ensuring semantic consistency, several attempts \cite{mo2024dynamic,hei2024user,zhan2024prompt} have been made in previous T2I works for prompt optimization. These efforts primarily involve instructing a pre-trained or fine-tuned Large Language Model (LLM) to incorporate detailed modifiers into original prompts, with the aim of enhancing spatial elements such as color and relationships. While these approaches have displayed promising outcomes in image generation, studies \cite{hao2024optimizing,chen2024cat} reveal that their impact on video generation remains limited, especially in terms of enhancing temporal aspects such as motion smoothness and minimizing temporal flickering. 

For T2V generation, we observe that the quality of spatial and temporal elements is significantly influenced by the selection of appropriate verb-object phrases and the structure of input prompts, which largely rely on analyzing and structuring the entire training data in a systematical manner. To this end, we propose \textbf{RAPO}, a \textbf{R}etrieval-\textbf{A}ugmented \textbf{P}rompt \textbf{O}ptimization framework for T2V generation. 

The primary objective of RAPO is to convert user-provided prompts into optimized prompts that retain the original semantics while closely adhering to the format of the training data, including the vocabulary and sentence structure. To achieve this, we initially construct a Relation Graph from the training data, with each node representing words and edges indicating the relationships between neighboring words. By leveraging any user-provided prompt, we can retrieve the most relevant terms to enhance the original descriptions.

Subsequently, a specially fine-tuned Refactoring LLM is introduced to reorganize the structure and writing style of the word-augmented prompt, aligning it more cohesively with the original training data. Additionally, we employ a Rewrite LLM to directly transform user-provided prompts into optimized prompts that follow a specific format consistent with the training prompts, serving as candidates for optimized prompts. In order to address potential inaccuracies and ambiguous details generated by LLMs, we conclude by fine-tuning a Discriminator LLM to identify the most suitable prompt for T2V generation.


We apply RAPO on two T2V generation models LaVie~\cite{wang2023lavie} and Latte~\cite{,ma2024latte}, and evaluate it on three benchmarks, namely VBench \cite{huang2024vbench}, EvalCrafter \cite{liu2024evalcrafter} and T2V-CompBench \cite{sun2024t2v}. It is worth noticing that RAPO greatly promotes the performance of multiple objects dimension in VBench from 37.71\% to 64.86\% using LaVie and from 29.55\% to 52.78\% using Latte. In addition, the experimental results show that RAPO improves the performance of T2V models to synthesize scenes with multiple subjects, even unusual scenes like \textit{"a mushroom growing out of a human head"} as shown in Fig.~\ref{fig:teaser}. Our contributions can be summarized as follows.

\begin{itemize}
    \item We propose RAPO, a Retrieval-Augmented Prompt Optimization framework T2V generation. 
    \item We propose word augmentation module and sentence refactoring module to optimize prompts considering prompt vocabulary and specific sentence format.
    \item We validate the effectiveness of RAPO through extensive experiments on several benchmarks. 
\end{itemize}

\section{Related Work}
\label{sec:Related Work}
\noindent \textbf{Text-to-Video generation.} With the remarkable breakthroughs  of diffusion models \cite{peebles2023scalable,ramesh2022hierarchical,rombach2022high},  the generations of 3D content \cite{yang2024layerpano3d, lin2023magic3d, chen2023fantasia3d}, images \cite{podell2023sdxl,esser2024scaling,betker2023improving}, and videos \cite{ho2022video,blattmann2023stable,brooks2024video} from text descriptions achieve rapid advancement. Text-to-Video (T2V) \cite{zhang2024show,chen2024videocrafter2,wang2023lavie} Generation aims to automatically create videos that match given textual descriptions. This process generally involves comprehending the scenes, objects, and actions described in the text and converting them into a sequence of cohesive visual frames, producing a video that is logically and visually consistent. T2V generation is wildly used in applications, such as animations \cite{he2023animate,guo2023animatediff,chen2023seine} and automatic movie generation \cite{zhao2024moviedreamer,zhuang2024vlogger,yang2023probabilistic}. However, large T2V generative models \cite{wang2023lavie,ma2024latte,yang2024cogvideox} trained on large-scale dataset could not adequately demonstrate their potential in generation due to mismatch between training and inference.

\noindent \textbf{Prompt optimization.} T2I and T2V generative models are sensitive to input prompts. However, the well-performed prompts are often model-specific and coherent with training prompts, misaligned with user input. Therefore, several studies \cite{hao2024optimizing,chen2024tailored,mo2024dynamic,zhan2024capability} are conducted to explore the generative potential of T2I and T2V generative models. Hao \emph{et al.} \cite{hao2024optimizing} propose a learning-based prompt optimizing framework unitizing reinforce learning for generating more aesthetically pleasing images. Chen \emph{et al.} \cite{chen2024tailored} enhance user prompts by leveraging the user's historical interactions with the system. Mo \emph{et al.} \cite{mo2024dynamic} propose Prompt Auto-Editing (PAE) method to decide the weights and injection time steps of each word without manual intervention. These methods primarily focus on prompts optimizing for T2I models and lack extension to T2V models. Yang \emph{et al.} \cite{yang2024cogvideox} use large language models (LLMs) to transform short prompts into more detailed ones, maintaining a consistent visual structure. Polyak  \emph{et al.} \cite{polyak2024movie} develop a teacher-student distillation approach for prompt optimization to improve computational efficiency and reduce latency. However, the results of optimized prompts usually could not be well-aligned with training prompts due to the misleading of the LLMs and the lack of more refined guidance.

\section{Method}
\label{sec:Method}
\begin{figure*}[tb]
    \centering
\includegraphics[width=1.0\textwidth]{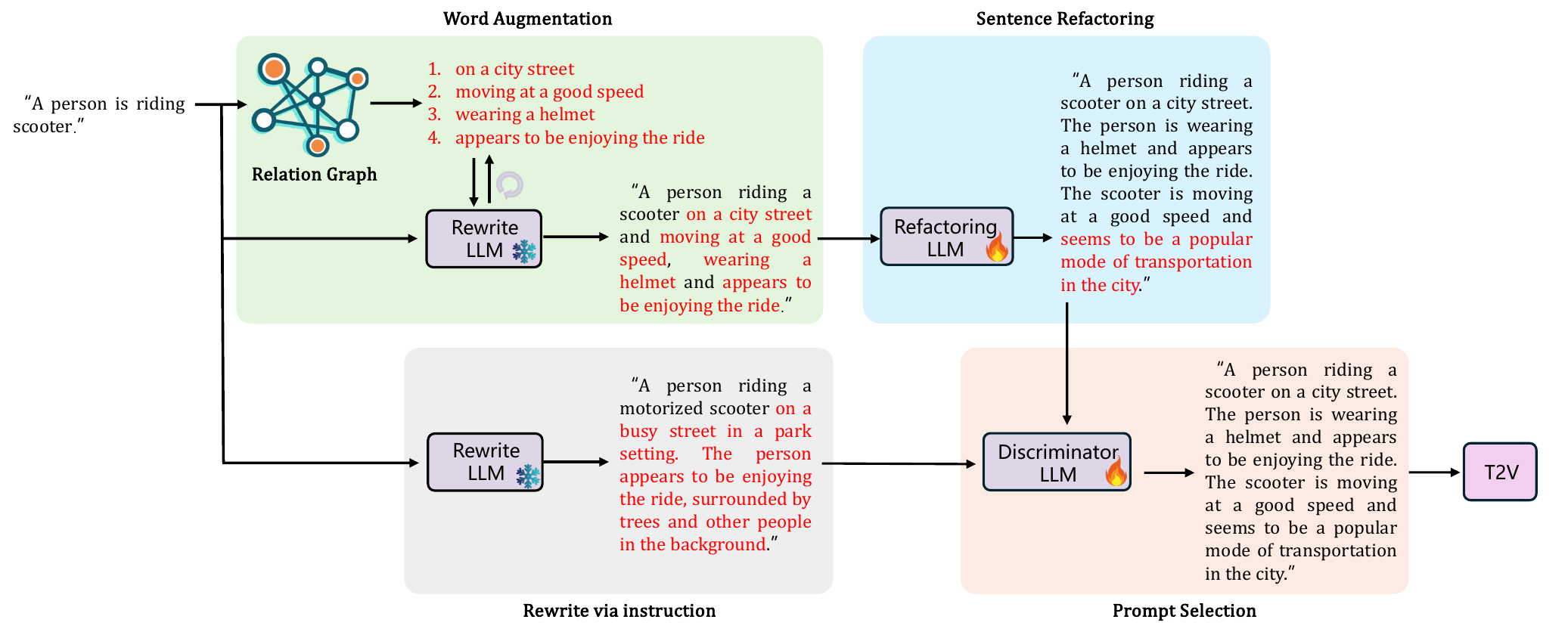}
    \caption{\textbf{Overview of RAPO.} The naive prompt is optimized by two branches respectively. For the first branch, it is enriched by the word augmentation module based on a constructed relation graph and a frozen Large Language Model (LLM). Subsequently, augmented prompt is refactored by a finetuned LLM into a specific format in the sentence refactoring module. For the second branch, the naive prompt is directly rewritten by a frozen LLM. Finally, the prompt selection module selects the better one from two branches' results as input for T2V model.}
    \label{fig:overview}
\end{figure*}
As illustrated in Fig.~\ref{fig:overview}, RAPO consists of three parts, 1) a \textit{word augmentation} module, 2) a \textit{sentence refactoring} module, as well as 3) a \textit{prompt selection} module. Given a user-provided prompt $x_i$, firstly, the word augmentation module utilizes an interactive retrieval-merge mechanism between a relation graph $\mathcal{G}$ and a LLM $\mathcal{L}$ to augment the prompt by adding related \textit{subject}, \textit{action} and \textit{atmosphere modifiers}. Then, a fine-tuned LLM $\mathcal{L}_r$ is applied to refactor the entire sentence into $x_r$. $x_r$ has a more unified format which is consistent with the prompt length and format distribution in training data. Finally, a discriminator in the prompt selection module decides between $x_r$ and a naively augmented prompt $x_n$ obtained directly from a LLM via instruction, as the most suitable augmented prompt for T2V generation. We proceed to introduce each module in detail in the following sections.

\subsection{Word Augmentation Module}
Given a user-provided prompt $x_i$, the word augmentation module aims to enrich $x_i$ with more multiple, straightforward  and relevant modifiers. It is achieved through retrieving modifiers meeting the requirements from a built relation graph $\mathcal{G}$, and merging them into $x_i$ through $\mathcal{L}$ via instruction. In this section, we first introduce the construction and retrieval of relation graph. And we introduce the instruction format and retrieval-merge mechanism of $\mathcal{L}$.

\begin{figure}[tb]
    \centering
\includegraphics[width=0.50\textwidth]{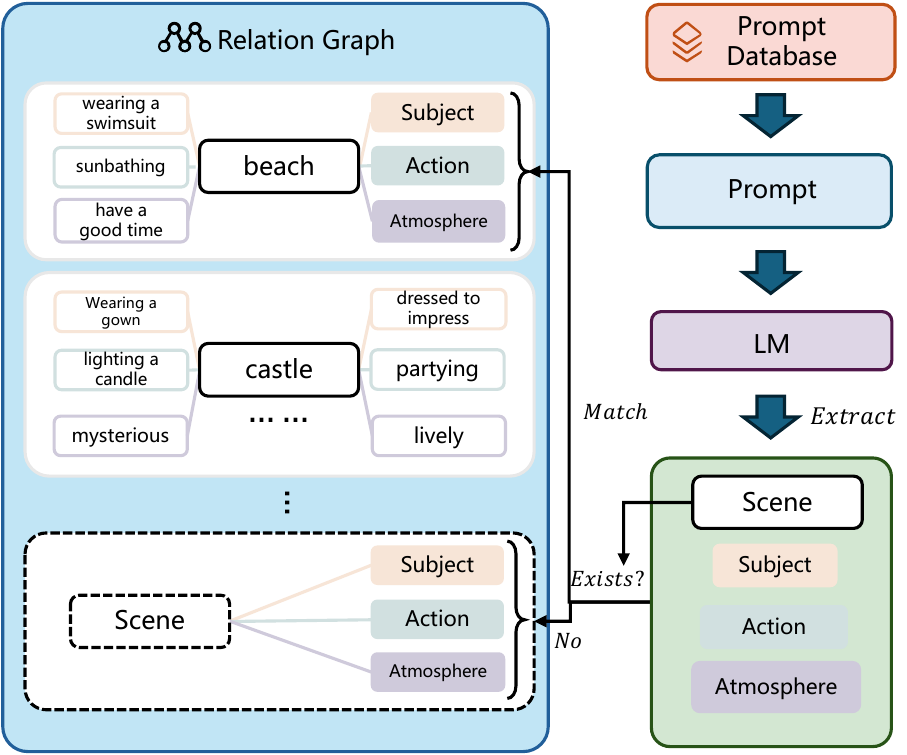}
    \caption{\textbf{The construction of relation graph.} Relation graph consists of multiple nodes (scenes acting as core nodes with modifiers connected as sub-nodes). For each prompt in database, LLM extracts scene and related modifiers. Based on whether the extracted scene is already in the graph or not, different methods are used to incorporate the new information into the graph.}
    \label{fig:relation_graph}
\end{figure}

\noindent \textbf{Relation Graph $\mathcal{G}$.} As shown in Fig.~\ref{fig:relation_graph}, we construct relation graph $\mathcal{G}$ based on training prompts database. For each training prompt, we utilize $\mathcal{L}$ to extract scene and corresponding related modifiers (subject, action, atmosphere descriptions). Each scene serves as a core node, with subject, action and atmosphere modifiers connected as individual sub-nodes in relation graph. For each extracted scene, we first check whether it exists in relation graph or not. If so the extracted related modifiers will be connected to the existing one. If not the extracted scene becomes a new core node with related modifiers connected. Finally, we can obtain a relation graph covering diverse scenes with multiple modifiers connected.

For relation graph retrieval, we utilize a sentence transformer pre-trained model to extract features of prompt, and employ the cosine similarity to measure similarity between sentence features. We first retrieve the top-k relevant scenes from $\mathcal{G}$ for $x_{i}$. Then we retrieve all modifiers connected to the retrieved scenes. We select the top-k relevant modifiers $\{p_{n}|_{n=0}^{k-1}\}$ from all retrieved modifiers, preparing for the retrieval-merge mechanism of $\mathcal{L}$.

\noindent \textbf{LLM $\mathcal{L}$.} We augment $x_i$ with retrieved modifiers $\{p_{n}|_{n=0}^{k-1}\}$ from relation graph. We rename $x_i$ with $x^0_i$ to illustrate the process of iterative merging. Specifically, the retrieved modifiers are merged into input prompt $x^0_i$ one by one through prompting $\mathcal{L}$, to maintain the information of the original input while adding relevant modifiers. 
\begin{equation}
x^{m+1}_{i} = f(x^{m}_{i},p^m_i),
\end{equation}
where $m=0,1,...,k-1$. $f$ is a function that combine $x^{m}_{i}$ and $p^m_i$ reasonably by $\mathcal{L}$. For instance, a merged prompt ``a woman dressed in a black suit representing a funeral" is resulted from merging the user-provided prompt "a woman representing a funeral" and a retrieved modifier "a black suit". We prompt $\mathcal{L}$ to perform general prompt merging in a normal manner as the template in Tab.~\ref{tab:Iterative Merging}. In instruction, we provide some prompt pairs $E=\{e_i|_{i=0}^{n}\}$ as examples, in which $e_i$ contains input prompt, a modifier and corresponding merged result.

\begin{table}[h]
\centering
\begin{tabular}{@{}p{0.95\columnwidth}@{}}
\toprule
\textbf{LLM Template for Retrieval-Merge Mechanism} \\ 
\midrule
Suppose you are a Text Merger. You receive two inputs from the user: a description body and a relevant modifier. Your task is to enrich the description body with relevant modifiers while retaining the description body. You should ensure that the output text is coherent, contextually relevant, and follows the same structure as the examples provided. \\
Examples of prompt-pairs provided: $E=\{e_i|_{i=0}^{n}\}$. \\
Input description body and modifier are: \{$x^m_i$,$p^m_i$\}. \\
The merged prompt is: \{$x^{m+1}_{i}$\}.\\
\bottomrule
\end{tabular}
\caption{\textbf{Input template for retrieval-merge mechanism.} This template specifies how a frozen LLM iteratively merges user-provided prompt texts with relevant modifiers retrieved from a relation graph, thereby enriching the prompt’s semantic content and aligning it with the training prompt structure for improved text-to-video synthesis.}
\label{tab:Iterative Merging}
\end{table}

\subsection{Sentence Refactoring Module}
\label{sec:Sentence-Level Refactoring}
Sentence refactoring module aims to refactor word augmented prompts from word augmentation module to be more consistent with prompt format in training data. It is achieved through a fine-tuned LLM $L_r$ named as refactoring model. In this section, we introduce the training data preparation and instruction tuning for $L_r$. 

\noindent \textbf{Data preparation.} We represent the required dataset for training refactoring model by $\{D_r=r_i|_{i=1}^{N^r}\}$, in which $r_i$ involves a pair of prompts and $N^r$ is the number of training prompts pairs. Specifically, $r_i=(w_{i},c_{i})$, in which $w_{i}$ targets to simulate world augmented prompt, and $c_{i}$ represents the target prompt, that is, a training prompt for T2V models. $w_{i}$ and $c_{i}$ share similar semantics while different in the prompt format and length. Therefore, we generate $w_{i}$ automatically through rewriting $c_{i}$ utilizing $\mathcal{L}$ via instruction to break the unified training prompt format but maintaining the original semantics. 

\noindent \textbf{Instruction tuning for $L_r$.} We employ instruction tuning for fine-tuning a LLM on our constructed dataset of instructional prompts and corresponding outputs. The constructed dataset is based on $\{D_r=r_i|_{i=1}^{N^r}\}$ containing instructional prompts and corresponding outputs. The template of the instruction tuning dataset for $L_r$ is shown as Tab.~\ref{tab:Sentence Refactoring Module}.

\begin{table}[h]
\centering
\begin{tabular}{@{}p{0.95\columnwidth}@{}}
\toprule
\textbf{Instruction Tuning Dataset for $L_r$} \\ 
\midrule
Instruction. Refine format and word length of the sentence: $w_{i}$. Maintain the original subject descriptions, actions, scene descriptions. Append additional straightforward actions to make the sentence more dynamic if necessary.  \\
Output: target prompt $c_{i}$. \\
\bottomrule
\end{tabular}
    \caption{\textbf{Instruction tuning dataset template for $L_r$.} This template directs LLM fine-tuning to restructure augmented prompts by adjusting their format while preserving semantics, aligning them with the training data’s style for improved T2V generation.}
\label{tab:Sentence Refactoring Module}
\end{table}

\subsection{Prompt Selection Module} 
As shown in Fig.~\ref{fig:overview}, prompt selection module contains a fine-tuned LLM $\mathcal{L}_d$ named prompt discriminator to select the better one between $x_r$ from sentence refactoring module, and a naively augmented prompt $x_n$ obtained directly from a LLM via instruction. In this section, we introduce the training data preparation and instruction tuning for $\mathcal{L}_d$. 

\noindent \textbf{Data preparation.} We represent the required dataset for training refactoring model by $\{D_d=d_i|_{i=1}^{N^d}\}$, in which $d_i$ contains three prompts and $N^d$ is the number of training prompts triples. Specifically, $d_i=(x_i,x_r,x_n,y_d)$, in which $y_d$ represents the discriminator label to select the better one for T2V generation from $x_r$ and $x_n$ given input prompt $x_i$. To simulate the user-provided prompts, we collect diverse prompts from several T2V benchmarks and generate more utilizing $\mathcal{L}$ via instruction. $x_r$ and $x_n$ can be obtained from the proposed RAPO as shown in Fig.~\ref{fig:overview} given $x_i$.  We determine $y_d$ through the evaluation of generated videos conditioned on $x_r$ and $x_n$. Specifically, the evaluations of T2V models performance involves diverse dimensions. For collected or generated prompts, we need to determine the evaluation dimension according to prompt content. We automatically decide the evaluation dimension of input prompts utilizing $\mathcal{L}$, then choose the corresponding metrics to evaluate generated videos.

\noindent \textbf{Instruction tuning for $L_d$.} Similar to $L_r$, we employ instruction tuning for $L_d$ based on $\{D_d=d_i|_{i=1}^{N^d}\}$. The template of the instruction tuning dataset for $L_d$ is shown as Tab.~\ref{tab:Prompt Selection Module}.
\begin{table}[h]
\centering
\begin{tabular}{@{}p{0.95\columnwidth}@{}}
\toprule
\textbf{Instruction Tuning Dataset for  $L_d$} \\ 
\midrule
Instruction. Given user-provided prompt $x_i$, select the better optimized prompt from $x_r$ and $x_n$. The chosen prompt is required to contain multiple, straightforward, and relevant modifiers about  $x_i$ while involving the semantics of $x_i$.\\
Output: discriminator label $y_d$. \\
\bottomrule
\end{tabular}
    \caption{\textbf{Instruction tuning dataset template for $L_d$.} This template aims to train a discriminator LLM that evaluates multiple refined prompts and selects the optimal one based on the inclusion of clear, straightforward modifiers and faithful semantic alignment.}
\label{tab:Prompt Selection Module}
\end{table}

\section{Experiments}
\begin{figure*}[tb]
    \centering
\includegraphics[width=1.0\textwidth]{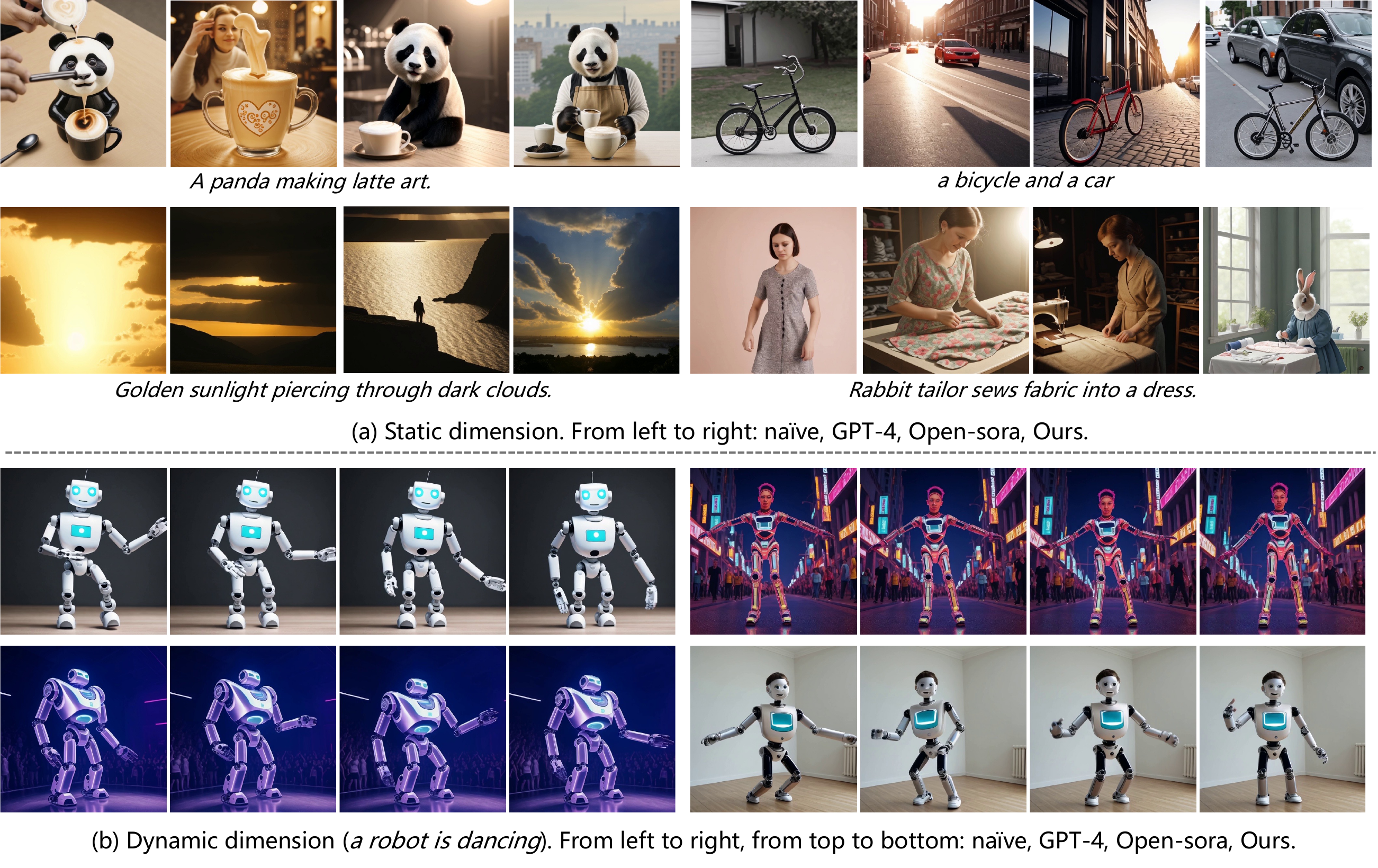}
    \caption{\textbf{Qualitative comparisons across dynamic and static dimensions.} This figure showcases videos generated using LaVie with short prompts, GPT-4 and Open-sora prompt optimizations, and our RAPO method. Videos produced with RAPO exhibit significantly sharper spatial details, smoother temporal transitions, and a closer semantic alignment with the input text.}
    \label{fig:results}
\end{figure*}
\subsection{Experimental Setup}
\noindent \textbf{Evaluation and Metrics.} We utilize the VBench \cite{huang2024vbench}, EvalCrafter \cite{liu2024evalcrafter} and T2V-CompBench \cite{sun2024t2v} for quantitative performance evaluation. VBench is an extensive benchmark comprising 16 fine-grained dimensions designed to systematically evaluate the quality of generated videos. EvalCrafter is a comprehensive evaluation benchmark that includes approximately 17 objective metrics for evaluating video generation performance. T2V-CompBench is the first
benchmark tailored for compositional text-to-video generation.

\noindent \textbf{Comparison to other methods.} We compare the optimized prompts with our method to three types of prompts: the short primary prompts, the prompts generated from GPT-4 \cite{achiam2023gpt} and prompt refiner from Open-sora \cite{open-sora-plan2024}, which is based on fine-tuned LLaMA 3.1 \cite{touvron2023llama}. The specific inference prompt template for GPT-4 \cite{achiam2023gpt} can be found in supplementary materials.

\subsection{Implementation Details} The well-performed prompts are model-specific and aligned with the distribution of training  prompts.  We employ Vimeo25M \cite{wang2023lavie}, a training dataset consisting of 25 million text-video pairs as our analysis dataset. At the same time, we choose LaVie \cite{wang2023lavie} and Latte \cite{ma2024latte} as analysis T2V models, which belong to the diffusion-based and DiT architectures respectively and use Vimeo25M as one of training datasets. And we represent some extension results to some other T2V models in supplementary materials. For relation graph construction, we utilize Mistral \cite{jiang2023mistral} to extract scenes with corresponding subject, action and atmosphere descriptions from Vimeo25M dataset, and use all-MiniLM-L6-v2 as sentence transformer pre-trained model. We filter about 2.1M valid sentences from from Vimeo25M dataset. For refactoring model training data, we prepare about 86k prompt-pairs following data preparation method in Section \ref{sec:Sentence-Level Refactoring}. For prompt discriminator training data, we first generate 7K text captions using Mistral, covering all the dimensions in VBench \cite{huang2024vbench}. The specific prompt template for extraction and dataset construction can be found in supplementary materials. We perform LoRA fine-tuning using LLaMA 3.1 \cite{touvron2023llama}, and fine-tune 8 epochs and 3 epochs for refactoring model and prompt discriminator respectively with a single A100, using a batch size of 32 and a LoRA rank of 64.


\begin{table*}[t]
  \centering
  \resizebox{\textwidth}{!}{ 
  \begin{tabular}{lccccccc}
    \toprule
     \textbf{Models} & \textbf{Total Score} &temporal flickering &imaging quality& human action & object class & multiple objects & spatial relationship   \\
    \midrule
    LaVie  & 80.89\%   &96.62\% &69.00\%  &95.80\%  &92.09\%  &37.71\%  &37.27\%  \\
    LaVie-GPT4  & 79.69\%  &96.14\% &70.27\%  &83.80\%  &88.73\%  &36.23\%  &50.55\%    \\
    LaVie-Open-sora  & 79.75\%   &96.42\%  & 70.42\% & 87.00\% &91.29\%  &36.52\%  &54.37\%  \\
    LaVie-Ours  & \textbf{82.38\%} &\textbf{96.86\%}   & \textbf{71.40\%} & \textbf{96.80\%}  & \textbf{96.91\%} &\textbf{64.86\%}  & \textbf{59.15\%}   \\
   Latte  & 77.03\%   & 97.10\% & 63.38\% & 88.40\%  & 83.86\% & 29.55\%  & 40.63\%  \\
   Latte-GPT4  & 77.40\%   & 97.52\%  & 63.54\%  & 85.80\%  & 78.32\%  & 27.73\%  & 36.72\% \\
    Latte-Open-sora  & 77.23\%   & 97.67\%  &64.19\%  & 84.60\% & 83.60\% & 30.00\% & 35.12\% \\
   Latte-Ours  & \textbf{79.97\%}  & \textbf{98.17\%} & \textbf{66.72\%} & \textbf{95.20\%} & \textbf{96.47\%} & \textbf{52.78\%} & \textbf{41.31\%} \\

  \bottomrule
  \end{tabular}
  }
\caption{\textbf{Quantitative comparisons on VBench.} RAPO achieves the highest overall scores, especially in static attributes and multiple object generation.}
\label{tab:Quanti-vbench}
\end{table*}

\subsection{Evaluation Results}
\noindent \textbf{Qualitative comparisons.}
As shown in Fig.~\ref{fig:results}, compared with user-provided prompts and other prompt optimization methods, RAPO centers the input semantics and promotes the the quality of generated videos from static and dynamic dimensions. The relevant modifiers promote model comprehending the simple user-provided prompts. The detailed prompts of different methods and more qualitative comparisons are provided in supplementary materials. 

\noindent \textbf{Quantitative comparisons.}
As shown in Tab.~\ref{tab:Quanti-vbench} and  Tab.~\ref{tab:Quanti-EvalCrafter}, the results show that RAPO surpasses other methods in static dimensions (\emph{e.g.},  visual quality, object class ) and dynamic dimensions (\emph{e.g.},  human actions, temporal flickering ). The consistent performance across various benchmarks demonstrates the robustness and versatility of RAPO. Although the optimized prompts from GPT-4 and Open-sora enriches the plain user-provided prompts with more details describing the scenes, objects, and actions, these lengthy and complex descriptions may confuse models and even worsen the generation. It is worth noticing that RAPO significantly enhances the score of multiple objects dimension from 37.71\% to 64.86\% using LaVie and from 29.55\% to 52.78\% using Latte as shown in Tab.~\ref{tab:Quanti-vbench}. And RAPO outperforms other methods on T2V-CompBench as shown in Tab.~\ref{tab:Quanti-T2V-CompBench} for some selected dimensions, which are focused on complex compositional T2V tasks. These experiments verified that PAPO significantly improves the performance of generating scenes involving more than two subjects.

\begin{table*}[tb]
  \centering
  \resizebox{\textwidth}{!}{ 
  \begin{tabular}{lccccc}
    \toprule
     \textbf{Models} & \textbf{Final Sum Score} & motion quality
& text-video alignment & visual quality & temporal consistency  \\
    \midrule
    LaVie &  248  & 53.19 & 69.60 &  64.81 & 60.87 \\
    LaVie-GPT4  & 246   & \textbf{54.05} &65.51  & 64.96 &61.22  \\
    LaVie-Open-sora  &251    &53.07  &71.38  &65.26  &\textbf{61.41}   \\
    LaVie-Ours  &  \textbf{256}  & 53.34  & \textbf{74.38} &\textbf{66.62}  & 61.29  \\
   Latte  & 217   & 50.03 & 55.49 &57.65  & 53.94  \\
   Latte-GPT4  & 218   & 51.36  & 53.65 & 58.02 &  54.65  \\
    Latte-Open-sora  & 222 & 50.25 & 57.32 & 58.71 &  \textbf{55.47}  \\
   Latte-Ours  & \textbf{227}   & \textbf{51.73} & \textbf{60.86} & \textbf{59.24}  &  55.26  \\

  \bottomrule
  \end{tabular}
  }
\caption{\textbf{Quantitative comparisons on EvalCrafter.} The results demonstrate that RAPO consistently achieves higher scores, especially in terms of text-video alignment and visual quality, indicating its superior ability to enhance the overall quality of generated videos.}
\label{tab:Quanti-EvalCrafter}
\end{table*}

\begin{table*}[tb]
  \centering
  \resizebox{\textwidth}{!}{ 
  \begin{tabular}{lccccc}
    \toprule
     \textbf{Models} & consistent attribute
binding &  dynamic attribute binding & action binding & object interactions \\
    \midrule
    LaVie   & 0.620 & 0.232  &0.483  &0.760 \\
    LaVie-GPT4      & 0.561 & 0.218 & 0.428 & 0.620  \\
    LaVie-Open-sora    & 0.532 & 0.214 &0.470  &  0.698  \\
    LaVie-Ours    &  \textbf{0.692} & \textbf{0.267} & \textbf{0.635} & \textbf{0.839}  \\
   Latte  & 0.633 & 0.227 & 0.476 &  0.792 \\
   Latte-GPT4 & 0.598  & 0.210  & 0.405 &  0.688   \\
    Latte-Open-sora   & 0.549 & 0.203 & 0.487 &  0.743  \\
   Latte-Ours   & \textbf{0.706}  & \textbf{0.258} & \textbf{0.591} &  \textbf{0.856}  \\

  \bottomrule
  \end{tabular}
  }
\caption{\textbf{Quantitative comparisons on T2V-CompBench.} RAPO outperforms alternative prompt optimization methods, exhibiting better performance at managing complex compositional tasks and ensuring coherent interactions between multiple objects in the generated videos.}
\label{tab:Quanti-T2V-CompBench}
\end{table*}

\subsection{Analyses}

\begin{figure}[tb]
    \centering
\includegraphics[width=0.5\textwidth]{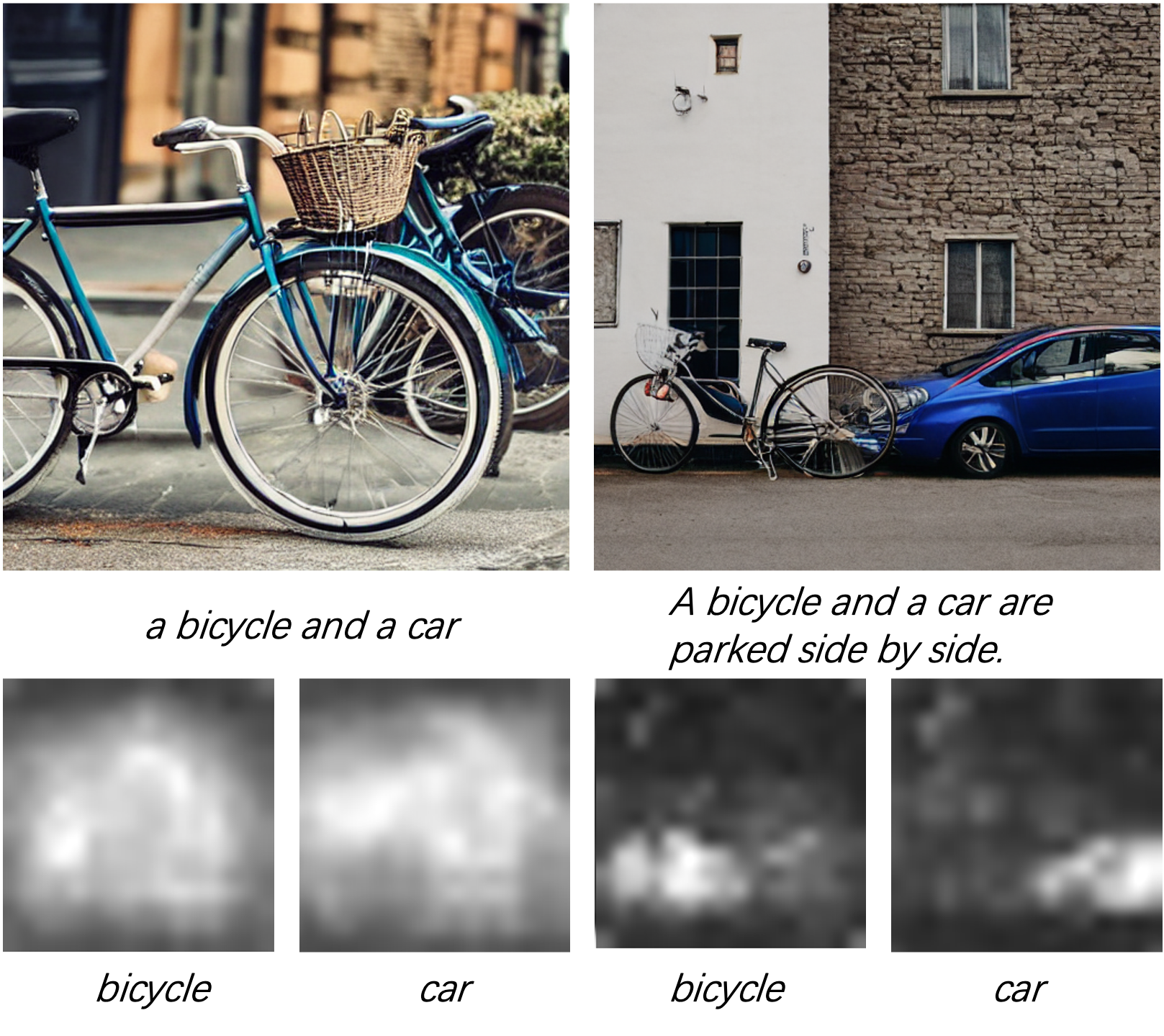}
    \caption{\textbf{Visualization on attention map on multiple objects from different prompts.} Adding description of the relative spatial position between objects can improve multi-object generation.}
    \label{fig:attn_map}
\end{figure}

\noindent \textbf{Multiple objects.} Synthesis quality of generated videos often declines when tasked with generating outputs that accurately represent prompts involving multiple objects. This issue is also prevalent in the
T2I model, and several studies \cite{feng2022training,podell2023sdxl,han2024aucseg} have highlighted that the blended context created by the CLIP text encoder leads to improper binding. Meanwhile, some related works \cite{chefer2023attend,phung2024grounded} focus on image latents to address information loss, while the others \cite{chen2024cat,zhuang2024magnet} pay more attention to text embedding to deal with the issue. However, few have explored optimizing prompts to improve the performance of multiple obejsts task. We apply our method to text-to-image using SD 1.4 \cite{rombach2022high}, which uses the same text encoder with LaVie \cite{wang2023lavie}. We test on prompts about multiple objects, and remove the irrelevant modifiers like action and atmosphere descriptions. As shown in Fig.~\ref{fig:attn_map}, we can find the relevant spatial descriptions boost the performance of multiple objects. More examples can be found in supplementary materials.

\noindent \textbf{Statistical analysis of text.}
\begin{figure}[tb]
    \centering
\includegraphics[width=0.45\textwidth]{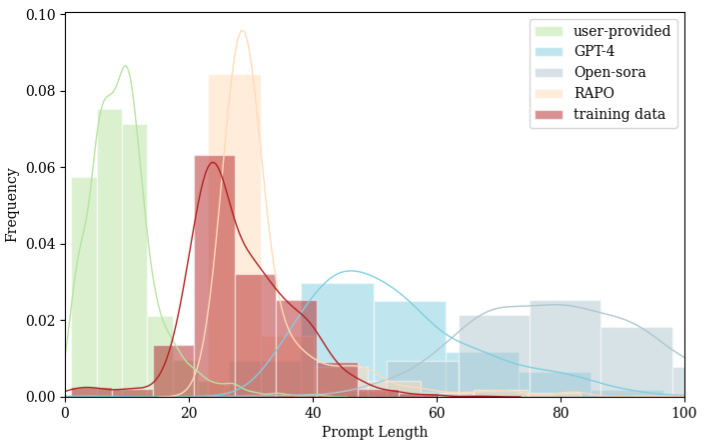}
    \caption{\textbf{Prompt length distribution comparison among various methods.} The distribution of RAPO-optimized prompts is more closer to the training prompts.}
    \label{fig: prompt_length}
\end{figure}
As shown in Fig.~\ref{fig: prompt_length}, we compared the word length distributions of prompts from the T2V training set, user prompts (simulated via VBench, EvalCrafter, and T2V-CompBench), and optimized prompts generated by various methods. The results show that the prompt length distribution produced by RAPO is closest to that of the training set, and this consistency unleashes the model's generative potential to produce better videos. In contrast, user prompts are too short and lack necessary details, while other methods generate longer prompts that contain excessive details and complex vocabulary, which may be counterproductive, as shown in Tab.~\ref{tab:Quanti-vbench} and Tab.~\ref{tab:Quanti-EvalCrafter}.

\subsection{Ablation Study}
We conduct ablation experiments on the VBench benchmark to examine the individual and combined effects of different modules in RAPO. Additionally, we perform ablation experiments on various configurations of $\mathcal{L}$. We present more ablation experiments on hyperparameters in the supplementary materials.

\noindent \textbf{Ablating each modules in RAPO.}  We directly obtain the related modifiers about input prompts utilizing GPT-4 \cite{achiam2023gpt}, and merging them into inputs at one time as the comparison of word augmentation. We randomly select one of optimized prompts as the comparison of prompt selection. The optimal result is achieved by the full-fledged framework as shown in row (f).
\begin{table}[t]
  \centering
  \resizebox{\columnwidth}{!}{ 
  \begin{tabular}{lccccc}
    \toprule
     &word augmentation&sentence refactoring&prompt selection&  VBench Total Score  \\
    \midrule
    (a)&\(\checkmark\) & &  &80.37\%  \\
    (b)&  &\(\checkmark\) &      &79.75\%  \\
    (c)&\(\checkmark\) &\(\checkmark\) &      &81.58\%  \\
    (d)& &\(\checkmark\) &\(\checkmark\)   & 81.75\%   \\
    (e)&\(\checkmark\) & &\(\checkmark\)     &80.60\% \\
    (f)&\(\checkmark\) &\(\checkmark\) &\(\checkmark\)      &\textbf{82.38\%} \\
  \bottomrule
  \end{tabular}
  }
    \caption{\textbf{Ablation studies of different modules in RAPO on VBench.} Each module improves performance, while the combined use of all three leads to the highest evaluation score, confirming the synergistic effect of the full RAPO framework.}
    \label{tab:ab1}
\end{table}

\begin{table}[t]
  \centering
  \resizebox{\columnwidth}{!}{ 
  \begin{tabular}{ccccc}
    \toprule
     &&GPT-4&Mistral&  LLaMA  \\
    \midrule
    &VBench Total Score &82.38\% & 82.25\% &82.10\%  \\
  \bottomrule
  \end{tabular}
  }
    \caption{\textbf{Ablation studies on different $\mathcal{L}$.} The results suggest that  RAPO is robust and effective across various LLMs.}
    \label{tab:ab2}
\end{table}

\noindent \textbf{Ablation experiments on different $\mathcal{L}$.} 
We conduct ablation experiments on GPT-4 \cite{achiam2023gpt}, Mistral \cite{jiang2023mistral} and LLaMA 3.1 \cite{touvron2023llama}. As shown in Tab.~\ref{tab:ab2}, although GPT-4 achieves the best overall score, the differences are marginal, which suggests that RAPO is robust and effective across various LLMs in generating optimized prompts for T2V generation.

\section{Conclusion and Future Work}
In this paper, we propose RAPO, a novel framework for prompt optimizing to improve the static and dynamic dimensions performance of generated videos by pre-trained T2V models. RAPO employs word augmentation, sentence refactoring and prompt selection to generate prompts aligned with the distribution of training prompts,  involving relevant and straightforward modifiers about inputs. Extensive experimental evaluations demonstrate the effectiveness and efficiency of RAPO. For the future work, we plan to extend RAPO to more T2V models and conduct more analyses about the reasons for improvements.


{
    \small
    \bibliographystyle{ieeenat_fullname}
    \bibliography{main}
}


\end{document}